\documentclass[10pt,twocolumn,letterpaper]{article}

\usepackage[pagenumbers]{cvpr} 

\usepackage[dvipsnames]{xcolor}

\definecolor{cvprblue}{rgb}{0.21,0.49,0.74}
\usepackage[pagebackref,breaklinks,colorlinks,citecolor=cvprblue]{hyperref}
\usepackage{multirow}
\usepackage{xcolor}
\usepackage{colortbl}
\usepackage{bbding}
\usepackage{setspace}

\def\confName{CVPR}
\def\confYear{2024}

\title{HiH: A Multi-modal Hierarchy in Hierarchy Network for \\ Unconstrained Gait Recognition}

\author{Lei Wang, Bo Liu\thanks{Corresponding Author\\This work has been submitted to the IEEE for possible publication. Copyright may be transferred without notice, after which this version may no longer be accessible.}, Yinchi Ma, Fangfang Liang, Nawei Guo \\
College of Information Science and Technology, Hebei Agricultural University, China
}

\begin{document}
\maketitle
\begin{abstract}

Gait recognition has achieved promising advances in controlled settings, yet it significantly struggles in unconstrained environments due to challenges such as view changes, occlusions, and varying walking speeds. Additionally, efforts to fuse multiple modalities often face limited improvements because of cross-modality incompatibility, particularly in outdoor scenarios. To address these issues, we present a multi-modal Hierarchy in Hierarchy network (HiH) that integrates silhouette and pose sequences for robust gait recognition. HiH features a main branch that utilizes Hierarchical Gait Decomposer (HGD) modules for depth-wise and intra-module hierarchical examination of general gait patterns from silhouette data. This approach captures motion hierarchies from overall body dynamics to detailed limb movements, facilitating the representation of gait attributes across multiple spatial resolutions. Complementing this, an auxiliary branch, based on 2D joint sequences, enriches the spatial and temporal aspects of gait analysis. It employs a Deformable Spatial Enhancement (DSE) module for pose-guided spatial attention and a Deformable Temporal Alignment (DTA) module for aligning motion dynamics through learned temporal offsets. Extensive evaluations across diverse indoor and outdoor datasets demonstrate HiH's state-of-the-art performance, affirming a well-balanced trade-off between accuracy and efficiency.

\end{abstract}    
\section{Introduction}
\label{sec:intro}
Gait recognition aims to identify individuals by analyzing their walking patterns and styles captured uncooperatively from a distance \cite{sepas2022deep}. Compared to fingerprint recognition, gait offers the advantage of being contactless \cite{wan2018survey}. In contrast with facial recognition, gait patterns are more robust against spoofing and better preserve privacy, as gait analysis relies on human silhouette and movement rather than detailed visual features \cite{filipi2022gait}. Owing to these merits, gait has emerged as a promising biometric approach for applications like video surveillance \cite{wang2003silhouette,bouchrika2011using}, healthcare \cite{ren2014user,sun2020gait}, and forensics \cite{seckiner2019forensic,macoveciuc2019forensic}.

\begin{figure}[tbp]
  \centering
   \includegraphics[width=1.0\linewidth]{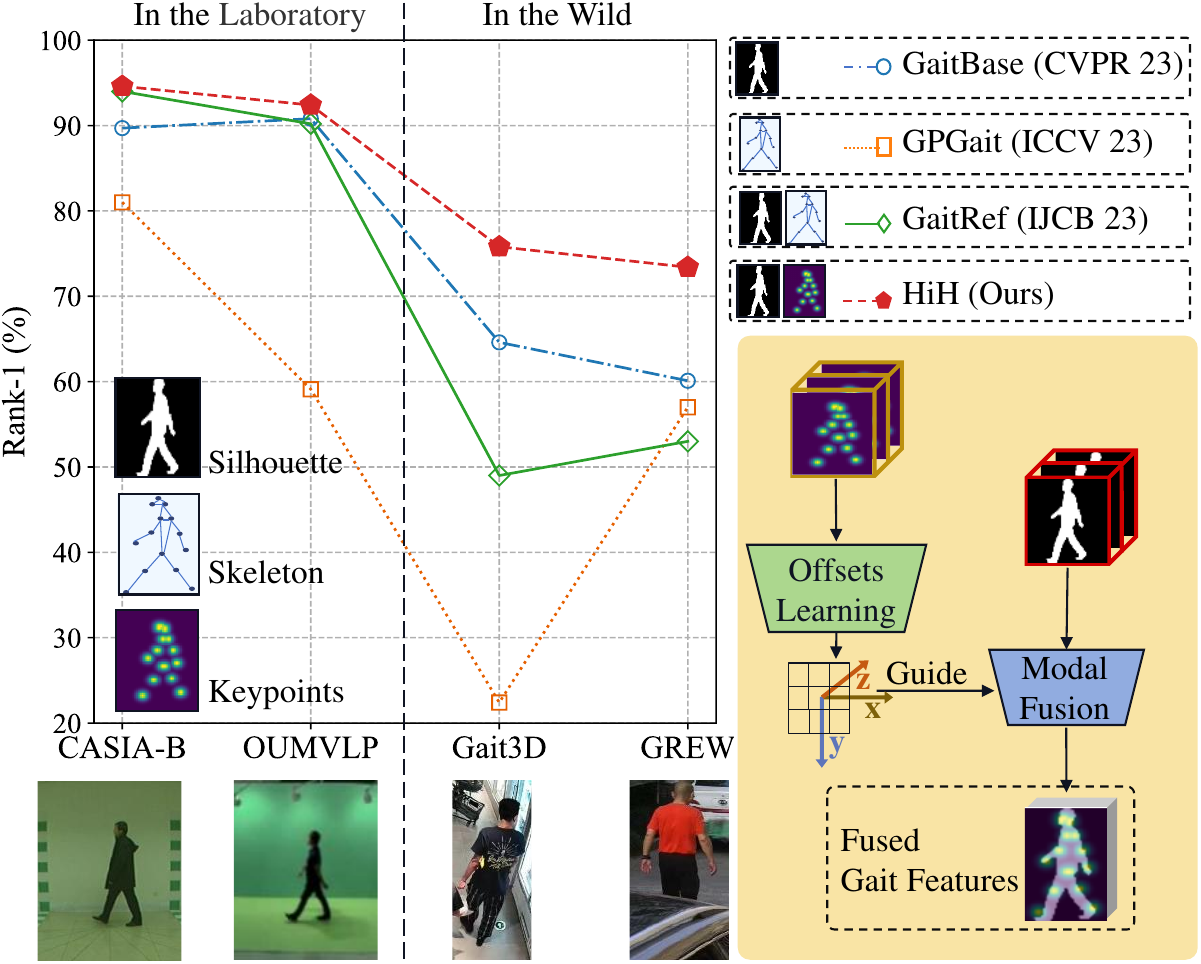}
   \caption{Motivation of the proposed HiH approach. Left: Performance degradation from controlled to uncontrolled scenarios. Right: Overview of HiH’s multi-modal fusion of silhouette and 2D keypoints sequences through pose-guided spatio-temporal processing.}
   \label{fig:motivation}
\end{figure}

In constrained or laboratory settings, existing gait recognition methods have achieved promising results. Particularly, appearance-based methods using binary silhouette images excel at capturing discriminative shape and contour information \cite{Ma_2023_CVPR,Dou_2023_CVPR,fan2023opengait,Wang_2023_ICCV,Wang_2023_ICCV_DyGait,wang2023gaitmm}. In parallel, model-based approaches explicitly estimate and exploit skeletal dynamics, uncovering view-invariant patterns robust to occlusions and cluttered backgrounds \cite{teepe2022towards,Fu_2023_ICCV}. By complementing silhouette information with pose data, multi-modal methods further enhance performance under challenging conditions like clothing and carrying variation \cite{peng2023learning,cui2023multi,hsu2023learning,zhu2023gaitref}. However, as the focus shifts towards in-the-wild scenarios to cater to real-world applications \cite{fan2023opengait, zheng2023parsing}, two main issues have emerged, as illustrated in the left of \cref{fig:motivation}. Firstly, algorithms highly effective in constrained settings often exhibit a significant decrease in performance when applied to outdoor benchmarks \cite{fan2023opengait,Fu_2023_ICCV,zhu2023gaitref}. This is attributed to covariates like camera view, occlusions, and step speed in real-world scenarios. Secondly, the incorporation of additional modalities like skeleton poses has not led to expected performance gains \cite{hsu2023learning,zhu2023gaitref}. The inherent data incompatibility between different modalities can introduce additional ambiguity.\looseness=-1

In light of the discussed challenges, we propose modeling two complementary aspects of gait: general gait motion patterns and dynamic pose changes, as depicted in the right of \cref{fig:motivation}. The general patterns, which can be extracted from silhouette sequences, refer to the kinematic gait hierarchy that manifests through biomechanics consistent across scenarios \cite{phinyomark2015kinematic,ferber2016gait}, thereby enhancing model generalization. Meanwhile, we employ 2D joints to represent evolving pose changes during walking. They circumvent potential inaccuracies of 3D skeleton estimation, especially in unconstrained settings. Moreover, mapping the 2D joints onto the image plane facilitates fusion with silhouette features and learning spatio-temporal offsets for deformation-based processing.

Specifically, we introduce a multi-modal Hierarchy in Hierarchy network (HiH) for unconstrained gait recognition. HiH consists of two branches. The main branch takes in silhouette sequences to model stable gait patterns. It centers on the Hierarchical Gait Decomposer (HGD) module and adopts a layered architecture via depth-wise and intra-module principles to unpack the kinematic hierarchy. In the depth-wise hierarchy, cascaded HGD modules progressively decompose motions into more localized actions across layers, enabling increasingly fine-grained feature learning. Meanwhile, the intra-module hierarchy in each HGD amalgamates multi-scale features to enrich global and local representations. Through joint modeling of the hierarchical structure across and within modules, the main branch effectively captures discriminative gait signatures. Complementing the main branch, the auxiliary branch leverages 2D pose sequences to enhance the spatial and temporal processing of the HGD modules in two ways: Spatially, the Deformable Spatial Enhancement (DSE) module highlights key local regions guided by the pose input. Temporally, the Deformable Temporal Alignment (DTA) module reduces redundant frames and extracts compact motion dynamics based on learned offsets. By providing pose cues, the auxiliary branch enhances the alignment of the main branch's learned representations with actual gait movements.

The main contributions are summarized as follows:
\begin{itemize}
    \item We propose the HiH network, a novel multi-modal framework for gait recognition in unconstrained environments. This network integrates silhouette and 2D pose data through the HGD, which executes a depth and width hierarchical decomposition specifically tailored to the complexities of gait analysis.
    \item We propose two pose-driven guidance mechanisms for HGD. DSE provides spatial attention to each frame using joints cues. DTA employs learned offsets to adaptively align silhouette sequences over time, reducing redundancy while adapting gait movement variations. 
    \item Comprehensive evaluation shows our HiH framework achieving state-of-the-art results on Gait3D and GREW in-the-wild and competitive performance on controlled datasets like OUMVLP and CASIA-B.This underscores its enhanced generalizability and a well-maintained balance between accuracy and efficiency.
\end{itemize}

\section{Related Work}
\label{sec:related_work}
\subsection{Single-modal Gait Recognition}
Single-modal gait recognition methods primarily leverage two modalities of input data: appearance-based modalities like silhouettes \cite{chao2019gaitset,fan2020gaitpart,Lin_2021_ICCV,dou2022metagait,zheng2022gaitmts,Dou_2023_CVPR,Wang_2023_ICCV} and model-based modalities like skeletons \cite{an2020performance,teepe2021gaitgraph,teepe2022towards,Fu_2023_ICCV} and 3D meshes \cite{Guo_2023_ICCV}. Appearance-based approaches directly extract gait features from raw input data. Earlier template-based methods such as the Gait Energy Image (GEI)~\cite{han2005individual} and Gait Entropy Image (GEnI)~\cite{bashir2009gait} create distinct gait templates by aggregating silhouette information over gait cycles.  These techniques compactly represent gait signatures while losing temporal details and being sensitive to viewpoint changes.

Recent silhouette-based methods have excelled by focusing on structural feature learning and temporal modeling. For structure, set-based methods like GaitSet \cite{chao2019gaitset} and Set Residual Network \cite{hou2021set} treat sequences as unordered sets, enhancing robustness to frame permutation. GaitPart \cite{fan2020gaitpart} emphasizes unique expressions of different body parts, with 3D Local CNN \cite{huang20213d} extracting part features variably. GaitGL \cite{Lin_2021_ICCV} and HSTL \cite{Wang_2023_ICCV} integrate local and global cues, though HSTL's pre-defined hierarchical body partitioning may limit its adaptability. Temporally, methods like Contextual relationships \cite{huang2021context}, second-order motion patterns \cite{chai2022lagrange}, and meta attention and pooling \cite{dou2022metagait} discern subtle patterns, with advanced techniques exploring dynamic mechanisms \cite{Ma_2023_CVPR, Wang_2023_ICCV_DyGait} and counterfactual intervention learning \cite{Dou_2023_CVPR} for robust spatio-temporal signatures. To harness the color and texture information in the original images, recent RGB-based gait recognition techniques aim to directly extract gait features from video frames, mitigating reliance on preprocessing like segmentation \cite{song2019gaitnet,zhang2019gait,li2021end,liang2022gaitedge}.  
\begin{figure*}[htbp]
  \centering
   \includegraphics[width=1.0\linewidth]{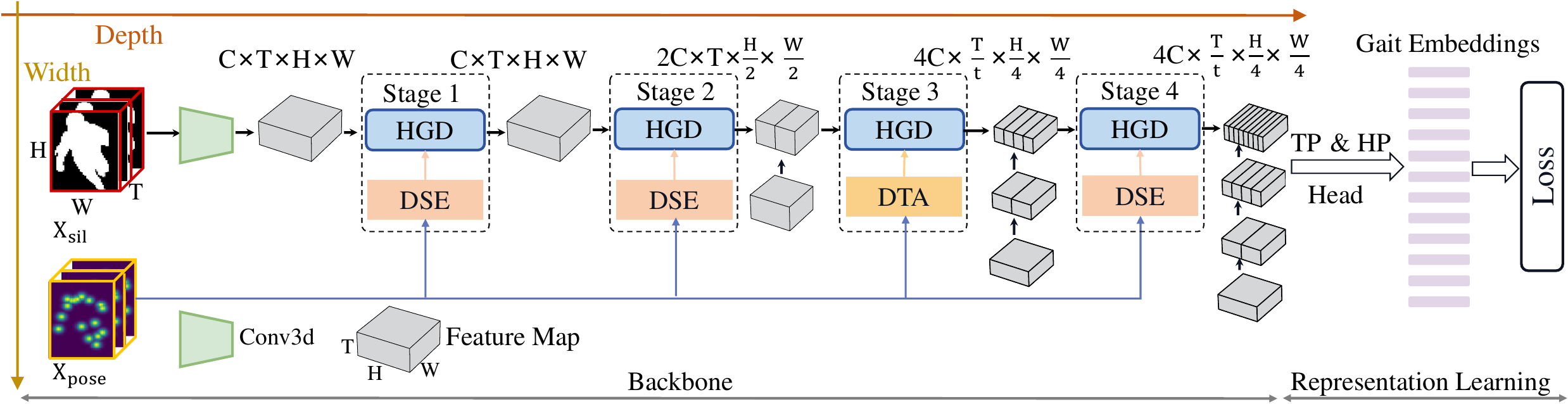}
   \caption{Overview of the HiH Framework. HiH takes silhouette sequence $X_{\text{sil}}$ and pose sequence $X_{\text{pose}}$ as inputs. The main branch uses multiple Hierarchical Gait Decomposers (HGDs) to  extract general gait motion patterns in both depth and width. The auxiliary branch enhances HGDs through pose-guided Deformable Spatial Enhancement (DSE) and  Deformable Temporal Alignment (DTA) modules, where DTA also performs temporal downsampling with stride $t$. Integrated outputs from both branches undergo Temporal Pooling (TP) and Horizontal Pooling (HP), and are then transformed into gait embeddings through fully-connected layers.}
   \label{fig:pipeline}
\end{figure*}
Model-based approaches build gait representations of body joints or 3D structure, then extract features and classify. Recent approaches utilize pose estimation advances to obtain cleaner skeleton input representing joint configurations\cite{teepe2021gaitgraph, Fu_2023_ICCV}. Graph convolutional networks help model inherent spatial-temporal patterns among joints \cite{Guo_2023_ICCV, teepe2022towards}. Some techniques incorporate biomechanical or physics priors to learn gait features aligned with human locomotion \cite{liao2020model, Guo_2023_ICCV}. Additionally, 3D mesh recovery from video has been explored for pose and shape modeling \cite{xu2023occlusion}.

\subsection{Multi-modal Gait Recognition}
Many recent approaches fuse complementary modalities like silhouette, 2D/3D pose, and skeleton to obtain more comprehensive gait representations. TransGait \cite{li2023transgait} combines silhouette appearance and pose dynamics via a set transformer model. SMPLGait \cite{zheng2022gait} introduces a dual-branch network leveraging estimated 3D body models to recover detailed shape and motion patterns lost in 2D projections. Other works focus on effective fusion techniques, including part-based alignment \cite{cui2023multi,peng2023learning,hsu2023learning} and refining skeleton with silhouette cues \cite{zhu2023gaitref}.  While fusing modalities like silhouette and pose has demonstrated performance gains in controlled settings, their effectiveness decreases on outdoor benchmarks. This is partly due to inaccurate skeleton pose estimation under unconstrained conditions, which causes difficulty in modality alignment. Unlike existing works, our approach utilizes more reliable 2D joint sequences to apply per-frame spatial-temporal attention correction to the silhouettes, achieving greater consistency across different modalities.

\section{Method}

In this section, we first overview the proposed Hierarchy in Hierarchy (HiH) framework (Sec. \ref{sec:framework_overview}). We introduce the Hierarchical Gait Decomposer (HGD) module for hierarchical gait feature learning (Sec. \ref{sec:HGD}), followed by the Spatially Enhanced HGD (Sec. \ref{sec:SE-HGD}) and Temporally Enhanced HGD (Sec. \ref{sec:TE-HGD}) modules, which strengthen HGD under the guidance of pose cues. Finally, we describe the loss function (Sec. \ref{sec:loss_function}).
\subsection{Framework Overview}
\label{sec:framework_overview}
The core of our proposed HiH framework integrates gait silhouettes with pose data to enhance gait recognition. As illustrated in \cref{fig:pipeline}, the framework operates on two input sequences: the silhouette sequence $X_{\text{sil}} \in \mathbb{R}^{C \times T \times H \times W}$ and the pose sequence $X_{\text{pose}} \in \mathbb{R}^{C \times T \times H \times W}$, where $C=1$ for both binary silhouette images and 2D keypoint-based pose representations. $T$ is the number of frames, and $H \times W$ denotes the spatial dimensions.

Building on the input sequences, the $\mathrm{HiH}$ framework utilizes a dual-branch backbone. The main branch processes $X_{\text {sil }}$ for general motion extraction. The initial step involves a 3D convolutional operation to extract foundational spatio-temporal features. This is followed by a series of stage-specific Hierarchical Gait Decomposer (HGD) modules, denoted as $\mathcal{H}_i$ for the $i$-th stage. The auxiliary branch leverages $X_{\text {pose }}$ to provide spatial and temporal guidance $\mathcal{G}\left(X_{\text {pose }}\right)$ to the HGD, via either Deformable Spatial Enhancement (DSE) or Deformable Temporal Alignment (DTA) modules. Thus, the output feature $F_i$ from each stage $\mathcal{H}_i$ is expressed as:
\begin{equation}
F_i=\mathcal{H}_i\left(X_{\text {sil }}, \mathcal{G}\left(X_{\text {pose }}\right)\right).
\end{equation}

Following the backbone, our framework applies Temporal Pooling (TP) \cite{Lin_2021_ICCV} and Horizontal Pooling (HP) \cite{fan2020gaitpart} to downsample the spatio-temporal dimensions. The reduced features are then processed through the head layer, which includes separate fully-connected layers and BNNeck \cite{luo2019bag}, effectively mapping them into a metric space. The model is optimized using separate triplet $\mathcal{L}_{\text{tri}}$ and cross-entropy $\mathcal{L}_{\text{ce}}$ losses.

\begin{figure}[t]
  \centering
   \includegraphics[width=0.75\linewidth]{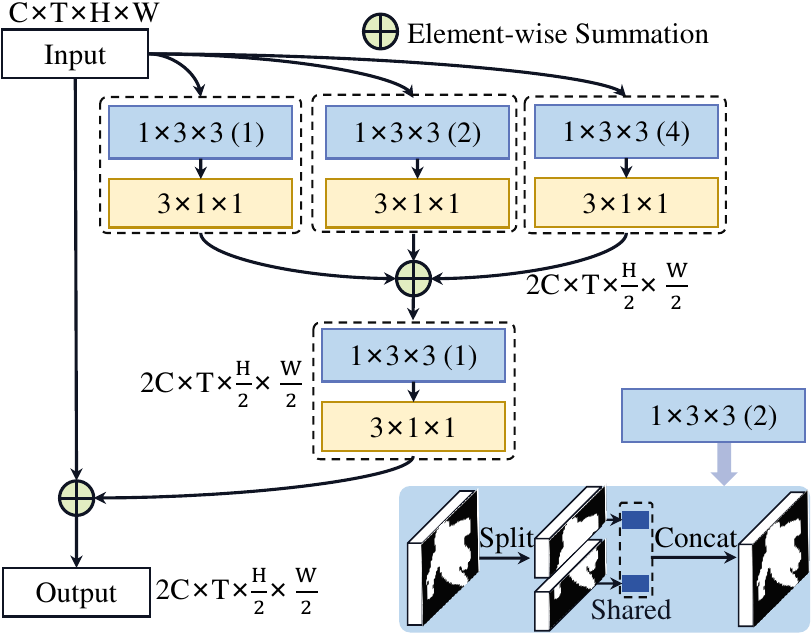}
   \caption{The architecture of the Hierarchical Gait Decomposer (HGD), where the number in parentheses denotes the amount of horizontal splits.}
   \label{fig:hgd}
\end{figure}

\subsection{Hierarchical Gait Decomposer (HGD)}
\label{sec:HGD}

Unlike part-based techniques that typically divide the body into uniform horizontal segments \cite{fan2020gaitpart,Lin_2021_ICCV}, the HGD employs a dual hierarchical approach for gait recognition, as shown in \cref{fig:pipeline}. This approach achieves a depth-wise hierarchy by stacking multiple HGD stages, which captures the gait dynamics from global body movements down to subtle limb articulations. In parallel, the width-wise hierarchy within each HGD stage conducts multi-scale processing to capture a comprehensive set of spatial features. The implementation of the $i$-th stage of the HGD, as illustrated in \cref{fig:hgd}, can be formalized as follows:

\begin{equation}
F_i^{\prime}=\sum_{n=1}^{2^{i-1}} f_{3 \times 1 \times 1}\left(f_{1 \times 3 \times 3}^{(n)}\left(F_{i-1}\right)\right),
\label{equ:res_hgd}
\end{equation}
where $f_{1 \times 3 \times 3}^{(n)}$ denotes the convolution operation  applied to $n$ horizontal strips of $F_{i-1}$ with a kernel size of $1 \times 3 \times 3$, designed to capture spatial features, while $f_{3 \times 1 \times 1}$ refines these features over the temporal dimension. Building on this multi-scale feature extraction, the aggregated features $F_i^{\prime}$ are further processed through a combination of additional convolutions and a residual connection \cite{he2016deep} by
\begin{equation}
F_i=f_{3 \times 1 \times 1}\left(f_{1 \times 3 \times 3}\left(F_i^{\prime}\right)\right)+F_{i-1}.
\end{equation}

\begin{figure}[t]
  \centering
   \includegraphics[width=0.82\linewidth]{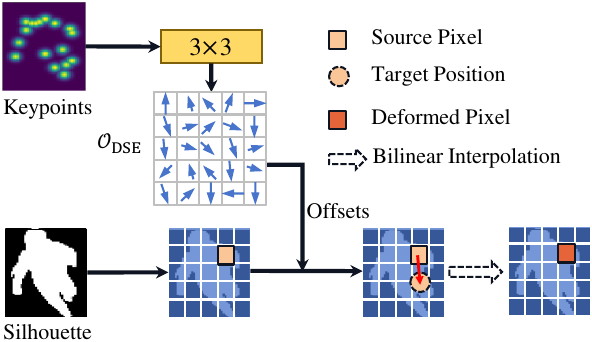}
   \caption{The detailed structure of the deformable spatial enhancement module (DSE).}
   \label{fig:dse}
\end{figure}

\subsection{Spatially Enhanced HGD (SE-HGD)} 
\label{sec:SE-HGD}

Silhouettes offer overall shape of gait descriptions but lack  structural details.  Fusing poses can provide complementary information on joint and limb movements. To achieve this, inspired by \cite{tu2022iwin,xia2022vision}, we introduce the Deformable Spatial Enhancement (DSE) module to adapt silhouettes using derived pose cues. As shown in \cref{fig:dse}, DSE utilizes learned deformable offsets to dynamically warp input silhouettes, emphasizing key spatial gait features and aligning them to corresponding poses. This forms a Spatially Enhanced HGD (SE-HGD) for more discriminative gait analysis.

Within the DSE, offsets are learned from pose input $X_{
\text{pose}}$ using a $3\times 3$ convolutional layer. The offsets are organized into a tensor $\mathcal{O}_{\text{DSE}} \in \mathbb{R}^{3 \times H \times W}$, prescribing spatial transformations for the silhouette input $X_{\text{sil}}$. $\mathcal{O}_{\text{DSE}}$ contains two components: offsets $\mathcal{O}^{x y}_{\text{DSE}}\in \mathbb{R}^{2 \times H \times W}$ representing pixel displacements in $x$ and $y$ directions, constrained by $\tanh$ activation, and offsets $\mathcal{O}_{\text{DSE-s}}^{xy}\in \mathbb{R}^{1\times H \times W}$ as scaling factors, processed via ReLU activation. The pixel-wise update to the silhouette input, utilizing the learned offsets, is conducted as follows:
\begin{equation}
X_{\text{DSE}}=\operatorname{BI}\left(X_{\text {sil }}, \tanh \left(\mathcal{O}_{\text{DSE}}^{x y}\right) \odot \operatorname{ReLU}\left(\mathcal{O}_{\text{DSE-s}}^{xy}\right)\right),
\end{equation}
where $\mathrm{BI}$ denotes the bilinear interpolation function that applies the spatial adjustments and scaling to $X_{\text {sil }}$, and $\odot$ represents element-wise multiplication.

\subsection{Temporally Enhanced HGD (TE-HGD)}
\label{sec:TE-HGD}

While silhouette sequences provide a outline of basic body movement, they typically fail to capture the intricate joint dynamics. Building on DSE, the proposed Deformable Temporal Alignment (DTA) module extends pose guidance to the temporal domain, adaptively aligning silhouettes to match gait variations. Combined with HGD, DTA constitutes the Temporally Enhanced HGD (TE-HGD). Moreover, DTA enables per-pixel sampling between frames. This makes temporal downsampling via fixed-stride max pooling more adaptive to motion variations. 



\begin{figure}[t]
  \centering
   \includegraphics[width=1.0\linewidth]{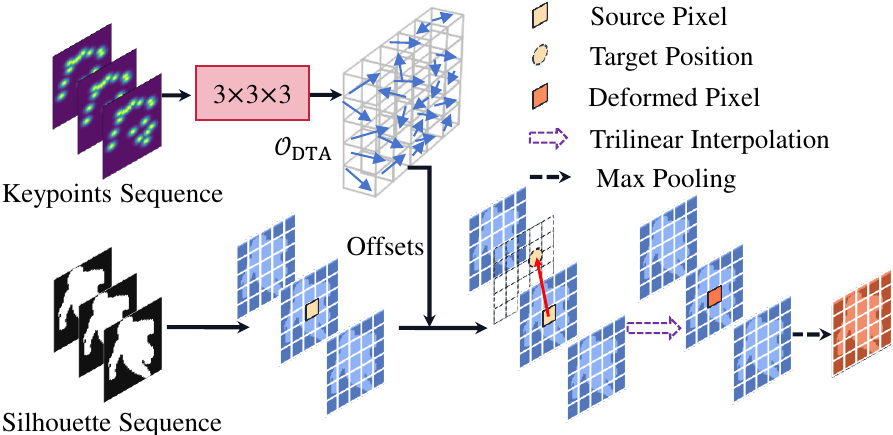}
   \caption{The detailed structure of the Deformable Temporal Alignment module (DTA).}
   \label{fig:dta}
\end{figure}
As illustrated in \cref{fig:dta}, the DTA module employs a 3D convolution $f_{3 \times 3 \times 3}$ on $X_{\text{pose}}$ to extract 5-channel spatio-temporal offsets $\mathcal{O}_{\text{DTA}} \in \mathbb{R}^{5\times T\times H\times W}$. The first two channels, $\mathcal{O}_{\text{DTA}}^{x y}$, capture spatial displacements in $x$ and $y$ axes, while the third, $\mathcal{O}_{\text{DTA}}^z$, quantifies temporal displacement. The fourth channel, $\mathcal{O}_{\text{DTA-s}}^{xy}$, scales spatial offsets, and the fifth, $\mathcal{O}_{\text{DTA-s}}^z$, adjusts temporal offsets. Similar to the DSE module,  $\mathcal{O}_{\text{DTA}}^{\prime}$ is formed by concatenating the processed offsets. This can be expressed as:
 \begin{equation}
     \begin{aligned}
\mathcal{O}_{\text{DTA}}^{\prime} = & \operatorname{concat}\left(\tanh \left(\mathcal{O}_{\text{DTA}}^{x y}\right) \odot \operatorname{ReLU}\left(\mathcal{O}_{\text{DTA-s}}^{x y}\right), \right. \\
& \hspace{1cm} \left. \tanh \left(\mathcal{O}_{\text{DTA}}^z\right) \odot \operatorname{ReLU}\left(\mathcal{O}_{\text{DTA-s}}^z\right)\right).
\end{aligned}
\end{equation}
These modified offsets $\mathcal{O}_{\mathrm{DTA}}^{\prime}$ guide the update of the silhouette features, followed by a MaxPooling operation to reduce redundancy. This process can be formulated as:
\begin{equation}
X_{\mathrm{DTA}}=\operatorname{MaxPool}\left(\operatorname{TI}\left(X_{\mathrm{sil}}, \mathcal{O}_{\mathrm{DTA}}^{\prime}\right)\right),
\end{equation}
where $\mathrm{TI}$ represents trilinear interpolation for spatio-temporal adjustments of the silhouette sequence, and $\text{MaxPool}$ is applied along the temporal dimension.

\subsection{Loss Function}
\label{sec:loss_function}
To effectively train our model, we use the joint losses  which include triplet loss $\mathcal{L}_{\text{tri}}$ \cite{hermans2017defense,Ma_2023_CVPR} and cross-entroy loss $\mathcal{L}_{\text{ce}}$ . The $\mathcal{L}_{\text{tri}}$ can be formulated as:
\begin{equation}
    \begin{gathered}
        \mathcal{L}_{\text{tri}}=\frac{1}{N_{\text{tri}}}\overbrace{\sum_{u=1}^U}^{\text{stripes}}\overbrace{\sum_{i=1}^P \sum_{a=1}^K}^{\text{anchors}}\overbrace{\sum_{\substack{p=1 \\ p \neq a}}^K}^{\text{positives}}\overbrace{\sum_{\substack{j=1, \\ j \ne i}}^P \sum_{n=1}^{K}}^{\text{negatives}}\left[m+ \right. \\ \left. d\left(\phi\left(x_{i,a}^{u}\right), \phi\left(x_{i,p}^{u}\right)\right)-d\left(\phi\left(x_{i,a}^{u}\right),\phi\left(x_{j,n}^{u}\right)\right)\right]_{+},
    \end{gathered}
    \label{equ:tri}
\end{equation}
where $N_\text{tri}$ represents the number of triplets with a positive loss, $U$ is the number of horizontal stripes, $P$ and $K$ are the number of subjects and sequences per subject, respectively, $\left[\gamma\right]_{+}$ equals to $\max\left(\gamma,0\right)$, $m$ is the margin, $d$ denotes the euclidean distance, and $\phi$ is the feature extraction function. The variables $x_{i, a}^u, x_{i, p}^u$, and $x_{j, n}^u$ represent the input sequences from anchors, positives, and negatives within the batch, respectively. 

The $\mathcal{L}_{\text{ce}}$ can be express as:
\begin{equation}
    \mathcal{L}_{\text{ce}}=-\frac{1}{P \times K}\overbrace{\sum_{i=1}^P \sum_{j=1}^K}^{\text{batch}}\overbrace{\sum_{n=1}^N}^{\text{subjects} }q_{i,j}^n\log\left(p_{i,j}^n\right),
    \label{equ:ce}
\end{equation}
where $N$ is the number of subject categories. In this formulation, $p_{i, j}^n$ denotes the predicted probability that the $j$-th sequence of the $i$-th subject in the batch belongs to the $n$-th category, and $q_{i, j}^n$ is the  ground truth label. Finally, combining \cref{equ:tri} and \cref{equ:ce}, the joint loss function $\mathcal{L}_{\text{joint}}$ can be formulated as:
\begin{equation}
\mathcal{L}_{\text{joint}}=\mathcal{L}_{\text{tri}}+\mathcal{L}_{\text{ce}}.
\end{equation}


\section{Experiments}
\subsection{Datasets}


We evaluated our method across four gait recognition datasets: Gait3D \cite{zheng2022gait} and GREW \cite{zhu2021gait}  from real-world environments, and OUMVLP \cite{takemura2018multi} and CASIA-B \cite{yu2006framework} from controlled laboratory settings .\looseness=-1

\noindent \textbf{Gait3D} \cite{zheng2022gait} is a large-scale wild gait dataset, featuring 4,000 subjects and 25,309 sequences collected from 39 cameras in a large supermarket. This dataset provides four modalities: silhouette, 2D and 3D poses, and 3D mesh. It is divided into a training set with 3,000 subjects and a test set comprising 1,000 subjects. During the evaluation phase, one sequence from each subject is randomly selected as the probe, while the remaining sequences form the gallery.


\noindent \textbf{GREW} \cite{zhu2021gait} is one of the largest wild gait datasets, containing 26,345 subjects and 128,671 sequences collected by 882 cameras. This dataset offers data in four modalities: silhouette, optical flow, 2D poses, and 3D poses. It is divided into training, validation, and test sets, containing 2,000, 345, and 6,000 subjects, respectively. For evaluation, two sequences from each subject in the test set are selected as probes, with the remaining sequences serving as the gallery. Additionally, GREW includes a distractor set with 233,857 unlabelled sequences.


\noindent \textbf{OUMVLP} \cite{takemura2018multi} is a large-scale indoor dataset, featuring 10,307 subjects captured from 14 different camera viewpoints. Each subject is recorded in two sequences under normal walking (NM) conditions. The dataset is evenly divided into a training set and a test set, containing 5,153 and 5,154 subjects, respectively. During the evaluation phase, sequences labeled NM\#01 are used as the gallery, while those labeled NM\#00 serve as the probe.


\noindent \textbf{CASIA-B} \cite{yu2006framework} is a widely-used indoor gait dataset that includes 124 subjects captured from 11 different viewpoints. Subjects were recorded under three walking conditions: normal walking (NM), walking with a bag (BG), and walking with a coat (CL). For evaluation purposes, we adopt the prevailing protocol, dividing the dataset into training and test sets with 74 and 50 subjects, respectively. During the evaluation, sequences NM\#01-04 are designated as the gallery, while the remaining sequences serve as probes. In addition, the HiH method requires precise frame alignment between RGB and silhouette sequences. However, we observed misalignment in the original CASIA-B dataset, hindering direct application. To address this, we utilize CASIA-B* \cite{liang2022gaitedge}, a variant with aligned RGB and silhouette data tailored to HiH's needs.



\subsection{Implementation Details}

\textbf{Training details.} 1) The margin $m$ in \cref{equ:tri} is set to 0.2, and the number of HP bins is 16;  2) Batch sizes are $(8,8)$ for CASIA-B, $(32,8)$ for OUMVLP, and $(32,4)$ for Gait3D and GREW; 3) Our input modalities include gait silhouettes and pose heatmaps generated by HRNet \cite{sun2019deep}, both resized to $64 \times 44$, with a fixed $\sigma$ of 2 for the 2D Gaussian distribution on keypoints. For training CASIA-B and OUMVLP,  30 frames are randomly sampled. For Gait3D, the frame range is $[10,50]$, and for GREW, it is $[20, 40]$, following \cite{fan2023opengait,Ma_2023_CVPR}; 4) The optimizer is SGD with a learning rate of 0.1 , training the model for 60K, 140K, 70K, and 200K iterations for CASIA-B, OUMVLP, Gait3D, and GREW, respectively; 5) In the training phase for two real-world datasets, data augmentation strategies (e.g., horizontal flipping, rotation, perspective) are applied as outlined in \cite{fan2023opengait,ma2023fine}.


\noindent\textbf{Architecture details.} 1) The output channels of the backbone in the four stages are set to $(64,64,128,256)$ for CASIA-B, $(64,128,256,256)$ for OUMVLP, Gait3D and GREW, respectively, to fit larger datasets; 2) Spatial downsampling is applied at the second and third stages for OUMVLP, Gait3D, and GREW, while not for CASIA-B; 3) For all four datasets, temporal downsampling with stride 3 is applied at the third stage.



\subsection{Comparison with State-of-the-Art Methods}
We conduct comprehensive comparisons between our proposed HiH method variants, HiH-S (using only silhouette modality) and HiH-M (integrating silhouettes with 2D keypoints for multimodal analysis), and three types of existing gait recognition methods: 1) Pose-based methods including GaitGraph2 \cite{teepe2022towards}, PAA \cite{Guo_2023_ICCV}, and GPGait \cite{Fu_2023_ICCV}; 2) Silhouette-based methods such as GaitSet \cite{chao2019gaitset}, GaitPart \cite{fan2020gaitpart}, GLN \cite{hou2020gait}, GaitGL \cite{Lin_2021_ICCV}, 3D Local \cite{huang20213d}, CSTL \cite{huang2021context}, LagrangeGait \cite{chai2022lagrange}, MetaGait \cite{dou2022metagait}, GaitBase \cite{fan2023opengait}, DANet \cite{Ma_2023_CVPR}, GaitGCI \cite{Dou_2023_CVPR}, STANet \cite{ma2023fine}, DyGait \cite{Wang_2023_ICCV_DyGait}, and HSTL \cite{Wang_2023_ICCV}; 3) Multimodal approaches like SMPLGait \cite{zheng2022gait}, TransGait \cite{li2023transgait}, BiFusion \cite{peng2023learning}, GaitTAKE \cite{hsu2023learning}, GaitRef \cite{zhu2023gaitref}, and MMGaitFormer \cite{cui2023multi}. \looseness=-1

\begin{table}[tp]
  \centering
  \caption{Rank-1 accuracy (\%), Rank-5 accuracy (\%), mAP (\%) and mINP (\%) on the Gait3D dataset.}
  \renewcommand{\arraystretch}{1}
  \resizebox{1.0\linewidth}{!}{
    \begin{tabular}{c|c|cccc}
    \toprule
    Method & Venue& Rank-1 & Rank-5 & mAP & mINP \\
    \midrule
    GaitGraph2 \cite{teepe2022towards}&CVPRW22 & 11.2  & -     & -     & - \\
    PAA \cite{Guo_2023_ICCV}  &ICCV23& 38.9  & 59.1  & -     & - \\
    GPGait \cite{Fu_2023_ICCV}&ICCV23 & 22.4  & -     & -     & - \\
    \midrule
    GaitSet \cite{chao2019gaitset}&AAAI19 & 36.7  & 58.3  & 30.0  & 17.3  \\
    GaitPart \cite{fan2020gaitpart} &CVPR20& 28.2  & 47.6  & 21.6  & 12.4  \\
    GLN \cite{hou2020gait}   &ECCV20& 31.4  & 52.9  & 24.7  & 13.6  \\
    GaitGL \cite{Lin_2021_ICCV} &ICCV21& 29.7  & 48.5  & 22.3  & 13.3  \\
    CSTL \cite{huang2021context} &ICCV21 & 11.7  & 19.2  & 5.6   & 2.6  \\
    
    GaitBase \cite{fan2023opengait} &CVPR23& 64.6  & -     & -     & - \\
    DANet \cite{Ma_2023_CVPR} &CVPR23& 48.0  & 69.7  & -     & - \\
    GaitGCI \cite{Dou_2023_CVPR} &CVPR23& 50.3  & 68.5  & 39.5  & 24.3 
    \\
   
    
    DyGait \cite{Wang_2023_ICCV_DyGait} &ICCV23& \underline{66.3}  & \underline{80.8}  & \underline{56.4}  & \underline{37.3}\\
    HSTL \cite{Wang_2023_ICCV}&ICCV23 & 61.3  & 76.3  & 55.5  & 34.8  \\
    \midrule
    \rowcolor{gray!30}
    HiH-S  &$-$& \textbf{72.4}  & \textbf{86.9}  & \textbf{64.4}  & \textbf{38.1}  \\
    \midrule
    SMPLGait \cite{zheng2022gait}&CVPR22 & 46.3  & 64.5  & 37.2  & 22.2  \\
    GaitRef \cite{zhu2023gaitref} &IJCB23& \underline{49.0}  & \underline{69.3}  & \underline{40.7}  & \underline{25.3}  \\
    \midrule
    
    \rowcolor{gray!30}
    HiH-M  &$-$& \textbf{75.8}  & \textbf{88.3}  & \textbf{67.3}  & \textbf{40.4}  \\
    \bottomrule
    \end{tabular}%
    }
  \label{tab:gait3d}%
\end{table}%

\begin{table}[tp]
  \centering
  \caption{Rank-1 accuracy (\%), Rank-5 accuracy (\%), Rank-10 accuracy (\%) and Rank-20 accuracy (\%) on the GREW dataset. }
  \renewcommand{\arraystretch}{1}
  \resizebox{1.0\linewidth}{!}{
    \begin{tabular}{c|c|cccc}
    \toprule
    Method & Venue& Rank-1 & Rank-5 & Rank-10 & Rank-20 \\
    \midrule
    GaitGraph2 \cite{teepe2022towards} &CVPRW22& 34.8  & -     & -     & - \\
    PAA \cite{Guo_2023_ICCV}  & ICCV23&38.7  & 62.1  & -     & - \\
    GPGait \cite{Fu_2023_ICCV} & ICCV23&57.0  & -     & -     & - \\
    \midrule
    GaitSet \cite{chao2019gaitset} &AAAI19& 46.3  & 63.6  & 70.3  & 76.8  \\
    GaitPart \cite{fan2020gaitpart} &CVPR20& 44.0  & 60.7  & 67.3  & 73.5  \\
    GaitGL \cite{Lin_2021_ICCV} &ICCV21& 47.3  & 63.6  & 69.3  & 74.2  \\
    CSTL \cite{huang2021context}&ICCV21 & 50.6  & 65.9  & 71.9  & 76.9  \\
    GaitBase \cite{fan2023opengait}&CVPR23 & 60.1  & -     & -     & - \\
    GaitGCI \cite{Dou_2023_CVPR}&CVPR23 & 68.5  & 80.8  & 84.9  & 87.7  \\
    STANet \cite{ma2023fine} &CVPR23& 41.3  & -     & -     & - \\
    DyGait \cite{Wang_2023_ICCV_DyGait} &ICCV23& \underline{71.4}  & \underline{83.2}  & \underline{86.8}  & \underline{89.5}  \\
    HSTL \cite{Wang_2023_ICCV} &ICCV23& 62.7  & 76.6  & 81.3  & 85.2  \\
    
    \midrule
    \rowcolor{gray!30}
    HiH-S  &$-$& \textbf{72.5}  & \textbf{83.6}  & \textbf{87.1}  & \textbf{90.0}  \\
    \midrule
    TransGait \cite{li2023transgait} &APIN23& \underline{56.3}  & \underline{72.7}  & \underline{78.1}  & \underline{82.5}  \\
    GaitTAKE \cite{hsu2023learning} &JSTSP23& 51.3  & 69.4  & 75.5  & 80.4  \\
    GaitRef \cite{zhu2023gaitref} &IJCB23& 53.0  & 67.9  & 73.0  & 77.5  \\
    
    
    
    \midrule
    \rowcolor{gray!30}
    HiH-M  &$-$& \textbf{73.4}  & \textbf{84.3}  & \textbf{87.8}  & \textbf{90.4}  \\
    \bottomrule
    \end{tabular}%
    }
  \label{tab:grew}%
\end{table}%

\noindent \textbf{Evaluation on Gait3D.} On the real-world Gait3D dataset, our method outperforms existing methods of both single-modality (pose-based, silhouette-based) and multi-modality methods across all metrics, as detailed in \cref{tab:gait3d}. Specifically, HiH-S achieves 6.1\%, 6.1\%, and 8.0\% higher Rank-1, Rank-5, and mAP than the state-of-the-art silhouette-based method DyGait, demonstrating the efficacy of dual-hierarchy modeling. HiH-M records 26.8\%, 19.0\%, and 26.6\% higher Rank-1, Rank-5, and mAP, respectively, than GaitRef.  Moreover, HiH-M achieves 3.4\% higher Rank-1 accuracy than HiH-S. These indicate that pose-guided learning supplements the fine details missing in silhouettes. It is observed that pose-based methods lag behind silhouette-based approaches, indicating that pose estimation in the wild remains challenging.


\begin{table*}[htbp]
  \centering
  \caption{Rank-1 accuracy (\%) on the OUMVLP dataset under all views, excluding identical-view cases. Std denotes the sample standard deviation of the performance across 14 different views.}
  \renewcommand{\arraystretch}{1}
  \resizebox{1.0\linewidth}{!}{
    \begin{tabular}{c|c|cccccccccccccc|c|c}
    \toprule
    \multirow{2}[4]{*}{Method} &\multirow{2}[4]{*}{Venue} & \multicolumn{14}{c|}{Probe View}                                                                              & \multirow{2}[4]{*}{Mean} & \multirow{2}[4]{*}{Std}\\
\cmidrule{3-16}  &        & $0^{\circ}$    & $15^{\circ}$   & $30^{\circ}$   & $45^{\circ}$   &$60^{\circ}$   & $75^{\circ}$   & $90^{\circ}$   & $180^{\circ}$  & $195^{\circ}$  & $210^{\circ}$  & $225^{\circ}$  & $240^{\circ}$  & $255^{\circ}$  & $270^{\circ}$  & \\
    \midrule
    GaitGraph2 \cite{teepe2022towards} &CVPRW22 & 54.3  & 68.4  & 76.1  & 76.8  & 71.5  & 75.0  & 70.1  & 52.2  & 60.6  & 57.8  & 73.2  & 67.8  & 70.8  & 65.3  & 67.1 &7.7 \\
    GPGait \cite{Fu_2023_ICCV} &ICCV23 &     $-$  &   $-$    &      $-$ &    $-$   &     $-$  &   $-$    & $-$      &    $-$   &     $-$  &   $-$    &     $-$  &    $-$  & $-$      &  $-$     & 59.1 &$-$ \\
    \midrule
    GaitSet \cite{chao2019gaitset}& AAAI19 & 79.3  & 87.9  & 90.0  & 90.1  & 88.0  & 88.7  & 87.7  & 81.8  & 86.5  & 89.0  & 89.2  & 87.2  & 87.6  & 86.2  & 87.1&4.0  \\
    GaitPart \cite{fan2020gaitpart}& CVPR20 & 82.6  & 88.9  & 90.8  & 91.0  & 89.7  & 89.9  & 89.5  & 85.2  & 88.1  & 90.0  & 90.1  & 89.0  & 89.1  & 88.2  & 88.7&2.3  \\
    GLN \cite{hou2020gait} & ECCV20  & 83.8  & 90.0  & 91.0  & 91.2  & 90.3  & 90.0  & 89.4  & 85.3  & 89.1  & 90.5  & 90.6  & 89.6  & 89.3  & 88.5  & 89.2 &2.1 \\
    CSTL \cite{huang2021context} & ICCV21 & 87.1  & 91.0  & 91.5  & 91.8  & 90.6  & 90.8  & 90.6  & 89.4  & 90.2  & 90.5  & 90.7  & 89.8  & 90.0  & 89.4  & 90.2 &1.1 \\
    GaitGL \cite{Lin_2021_ICCV}& ICCV21 & 84.9  & 90.2  & 91.1  & 91.5  & 91.1  & 90.8  & 90.3  & 88.5  & 88.6  & 90.3  & 90.4  & 89.6  & 89.5  & 88.8  & 89.7 &1.7 \\
    3D Local \cite{huang20213d}& ICCV21 & 86.1  & 91.2  & 92.6  & 92.9  & 92.2  & 91.3  & 91.1  & 86.9  & 90.8  & \textbf{92.2}  & \underline{92.3}  & 91.3  & 91.1  & 90.2  & 90.9 &2.0 \\
    LagrangeGait \cite{chai2022lagrange}& CVPR22& 85.9  & 90.6  & 91.3  & 91.5  & 91.2  & 91.0  & 90.6  & 88.9  & 89.2  & 90.5  & 90.6  & 89.9  & 89.8  & 89.2  & 90.0 &1.4 \\
    MetaGait \cite{dou2022metagait}&ECCV22 & 88.2  & 92.3  & \textbf{93.0}  & \textbf{93.5}  & \underline{93.1}  & \textbf{92.7}  & \textbf{92.6}  & 89.3  & 91.2  & 92.0  & \textbf{92.6}  & \underline{92.3}  & \textbf{91.9}  & 91.1  & 91.9  &1.4\\
    GaitBase \cite{fan2023opengait}& CVPR23&     $-$  &   $-$    &      $-$ &    $-$   &     $-$  &   $-$    & $-$      &    $-$   &     $-$  &   $-$    &     $-$  &    $-$  & $-$      &  $-$     &  90.8 &$-$ \\
    DANet \cite{Ma_2023_CVPR}&CVPR23 & 87.7  & 91.3  & 91.6  & 91.8  & 91.7  & 91.4  & 91.1  & 90.4  & 90.3  & 90.7  & 90.9  & 90.5  & 90.3  & 89.9  & 90.7& 1.0 \\
    
    GaitGCI \cite{Dou_2023_CVPR}& CVPR23& 91.2  & 92.3  & 92.6  & 92.7  & 93.0  & 92.3  & 92.1  & 92.0  & 91.8  & 91.9  & \textbf{92.6}  & 92.3  & 91.4  & 91.6  & 92.1 &\underline{0.5} \\
    STANet \cite{ma2023fine} &ICCV23 & 87.7  & 91.4  & 91.6  & 91.9  & 91.6  & 91.4  & 91.2  & 90.4  & 90.3  & 90.8  & 91.0  & 90.5  & 90.3  & 90.1  & 90.7 & 1.0\\
    HSTL \cite{Wang_2023_ICCV}& ICCV23 & \underline{91.4}  & \underline{92.9}  & \underline{92.7}  & \underline{93.0}  & 92.9  & \underline{92.5}  & \underline{92.5}  & \underline{92.7}  & \underline{92.3}  & \underline{92.1}  & \underline{92.3}  & 92.2  & \underline{91.8}  & \underline{91.8}  & \textbf{92.4} &\underline{0.5} \\
    \midrule
    
    BiFusion \cite{peng2023learning}&MTA23 & 86.2  & 90.6  & 91.3  & 91.6  & 90.9  & 90.8  & 90.5  & 87.8  & 89.5  & 90.4  & 90.7  & 90.0  & 89.8  & 89.3  & 89.9 & 1.4\\
    GaitTAKE \cite{hsu2023learning}&JSTSP23 & 87.5  & 91.0  & 91.5  & 91.8  & 91.4  & 91.1  & 90.8  & 90.2  & 89.7  & 90.5  & 90.7  & 90.3  & 90.0  & 89.5  & 90.4 &1.0 \\
    GaitRef \cite{zhu2023gaitref}& IJCB23& 85.7  & 90.5  & 91.6  & 91.9  & 91.3  & 91.3  & 90.9  & 89.3  & 89.0  & 90.8  & 90.8  & 90.1  & 90.1  & 89.5  & 90.2 & 1.5\\
    MMGaitFormer \cite{cui2023multi}&CVPR23 &     $-$  &   $-$    &      $-$ &    $-$   &     $-$  &   $-$    & $-$      &    $-$   &     $-$  &   $-$    &     $-$  &    $-$  & $-$      &  $-$     &  90.1 & $-$\\
    \midrule
    \rowcolor{gray!30}
    HiH-S & $-$& \textbf{92.1}  & \textbf{93.0}  & 92.4  & 92.7  & \textbf{93.2}  & \underline{92.5}  & 92.4  & \textbf{93.0}  & \textbf{92.4}  & 91.9  & 92.1  & \textbf{92.5}  & \textbf{91.9}  & \textbf{91.9}  & \textbf{92.4} & \textbf{0.4}\\
    \bottomrule
    \end{tabular}%
    }
  \label{tab:oumvlp}%
\end{table*}%

\noindent \textbf{Evaluation on GREW.} As shown in  \cref{tab:grew}, the results on GREW follow a similar trend as Gait3D. Even using only a single modality, our HiH-S achieves the best results among the methods compared, and HiH-M further improves Rank-1 accuracy by 0.9\% through multi-modality fusion. Our HiH-M addresses this issue by using 2D pose as an auxiliary modality, which retains essential gait information and avoids 3D joint motion errors.




\noindent \textbf{Evaluation on OUMVLP.} Since OUMVLP only provides silhouettes, we compare single-modality results. As \cref{tab:oumvlp} shows, HiH-S leads in mean Rank-1 accuracy. Notably, HiH-S excels in 8 of 14 camera views, particularly in those views like the front $0^{\circ}$ and back $180^{\circ}$ where the gait posture is less visible. Although the average result is on par with HSTL, our lower std suggests better cross-view stability while using less than half of HSTL's parameters (refer to \cref{sec:trade-off} for details). 

\begin{table}[tbp]
  \centering
  \caption{Rank-1 accuracy (\%) on the CASIA-B dataset under different walking conditions, excluding identical-view cases.}
  \renewcommand{\arraystretch}{1}
  \resizebox{1.0\linewidth}{!}{
    \begin{tabular}{c|c|ccc|c}
    \toprule
    Method & Venue & NM    & BG    & CL    & Mean \\
    \midrule
    GaitGraph2 \cite{teepe2022towards} & CVPRW22 & 80.3  & 71.4  & 63.8  & 71.8  \\
    GPGait \cite{Fu_2023_ICCV} & ICCV23 & 93.6  & 80.2  & 69.3  & 81.0  \\
    \midrule
    GaitSet \cite{chao2019gaitset} & AAAI19 & 95.0  & 87.2  & 70.4  & 84.2  \\
    GaitPart \cite{fan2020gaitpart} & CVPR20 & 96.2  & 91.5  & 78.7  & 88.8  \\
    GLN \cite{hou2020gait}   & ECCV20 & 96.9  & 94.0  & 77.5  & 89.5  \\
    CSTL \cite{huang2021context} & ICCV21 & 97.8  & 93.6  & 84.2  & 91.9  \\
    3D Local \cite{huang20213d} & ICCV21 & 97.5  & 94.3  & 83.7  & 91.8  \\
    GaitGL \cite{Lin_2021_ICCV} & ICCV21 & 97.4  & 94.5  & 83.6  & 91.8  \\
    LagrangeGait \cite{chai2022lagrange} & CVPR 22 & 96.9  & 93.5  & 86.5  & 92.3  \\
    MetaGait \cite{dou2022metagait} & ECCV22 & 98.1  & 95.2  & 86.9  & 93.4  \\
    DANet \cite{Ma_2023_CVPR} & CVPR23 & 98.0  & 95.9  & \textbf{89.9}  & \textbf{94.6}  \\
    GaitBase \cite{fan2023opengait} & CVPR23 & 97.6  & 94.0  & 77.4  & 89.7  \\
    GaitGCI \cite{Dou_2023_CVPR} & CVPR23 & \textbf{98.4}  & \textbf{96.6}  & 88.5  & \underline{94.5}  \\
    STANet \cite{ma2023fine} & ICCV23 & 98.1  & 96.0  & \underline{89.7}  & \textbf{94.6}  \\
    DyGait \cite{Wang_2023_ICCV_DyGait} & ICCV23 & \textbf{98.4}  & 96.2  & 87.8  & 94.1  \\
    HSTL \cite{Wang_2023_ICCV}  & ICCV23 & 98.1  & 95.9  & 88.9  & 94.3  \\
    \midrule
    \rowcolor{gray!30}
    HiH-S  & $-$ & \underline{98.2}  & \underline{96.3}  & 89.2  & \textbf{94.6}  \\
    \midrule
    TransGait \cite{li2023transgait}  & APNI23 & 98.1  & 94.9  & 85.8  & 92.9  \\
    BiFusion \cite{peng2023learning}  & MTA23 & \textbf{98.7}  & \underline{96.0}  & 92.1  & 95.6  \\
    GaitTAKE \cite{hsu2023learning}  & JSTSP23 & 98.0  & \textbf{97.5}  & 92.2  & \underline{95.9}  \\
    GaitRef \cite{zhu2023gaitref}  & IJCB23 & 98.1  & 95.9  & 88.0  & 94.0  \\
    MMGaitFormer \cite{cui2023multi}  & CVPR23 & \underline{98.4}  & \underline{96.0}  & \textbf{94.8}  & \textbf{96.4}  \\
    \bottomrule
    \end{tabular}%
    }
  \label{tab:casia-b}%
\end{table}%
\begin{table}[htbp]
  \centering
  \caption{Rank-1 accuracy (\%) on the CASIA-B* dataset under different walking conditions, excluding identical-view cases.}
  \footnotesize
  \renewcommand{\arraystretch}{1.0}
    \begin{tabular}{c|c|ccc|c}
    \toprule
    Method & Venue & NM    & BG    & CL    & Mean \\
    \midrule
    GaitSet \cite{chao2019gaitset} & AAAI19 & 92.3  & 86.1  & 73.4  & 83.9  \\
    GaitPart \cite{fan2020gaitpart} & CVPR20 & 93.1  & 86.0  & 75.1  & 84.7  \\
    GaitGL \cite{Lin_2021_ICCV} & ICCV21 & 94.2  & 90.0  & 81.4  & 88.5  \\
    GaitBase \cite{fan2020gaitpart} & CVPR23 & \underline{96.5}  & \underline{91.5}  & 78.0  & \underline{88.7}  \\
    \midrule
    \rowcolor{gray!30}
    HiH-S & $-$     & 94.6  & 91.1  & \underline{84.2}  & \underline{90.0}  \\
    \rowcolor{gray!30}
    HiH-M & $-$ & \textbf{96.8}  & \textbf{93.9}  & \textbf{87.0}  & \textbf{92.6}  \\
    \midrule
    \end{tabular}%
  \label{tab:casiab-star}%
\end{table}%
\noindent \textbf{Evaluation on CASIA-B.} Since the RGB videos and silhouette sequences in CASIA-B are not frame-aligned, we report HiH-S results for fair comparison. Other multimodal methods like MMGaitFormer mainly adopt late fusion and thus do not require frame alignment. As shown in \cref{tab:casia-b}, HiH-S achieves the highest average accuracy among single-modality methods, on par with top models like DANet and STANet. Moreover, HiH-S demonstrates strong generalizability, surpassing DANet by 24.4\% on Gait3D (see \cref{tab:gait3d}) and STANet by 31.2\% in Rank-1 on GREW (see \cref{tab:grew}). Benefiting from accurate pose estimation, multimodal methods show advantages on CASIA-B, especially for the CL condition. However, their performance degrades significantly on larger datasets like OUMVLP (see \cref{tab:oumvlp}) and more complex scenarios like Gait3D and GREW (see \cref{tab:grew,tab:gait3d}) , falling behind even single-modality methods. This highlights the strength of HiH in unconstrained settings. Moreover, to validate the effectiveness of HiH-M in indoor settings against other silhouette-based methods, we report results on CASIA-B* with aligned RGB and silhouettes in \cref{tab:casiab-star}.  HiH-M demonstrates the best performance, improving HiH-S by 2.2\%, 2.8\% and 2.8\% for three walking conditions, respectively. For more comparison results, such as cross-dataset evaluations, please refer to the supplementary materials.

\subsection{Ablation Study}
To validate the efficacy of each component in HiH, including HGD which provides hierarchical feature learning in depth and width, DSE and DTA for pose-guided spatial-temporal modeling, we conduct ablation studies the Gait3D dataset with results in \cref{tab:ablation}. The hierarchical depth and width provided by the HGD module lay a foundational baseline for our approach. Adding either DSE or DTA can further improve over this baseline. The best results are achieved when all modules are considered together.

\begin{table}[htbp]
  \centering
  \caption{Ablation study on the effectiveness of HGD, DSE, and DTA modules on the Gait3D dataset.}
  \renewcommand{\arraystretch}{0.8}
  \resizebox{1.0\linewidth}{!}{
    \begin{tabular}{cccc|cccc}
    \toprule
    \multicolumn{2}{c}{HGD} & \multirow{2}[4]{*}{DSE} & \multirow{2}[4]{*}{DTA} & \multirow{2}[4]{*}{Rank-1} & \multirow{2}[4]{*}{Rank-5} & \multirow{2}[4]{*}{mAP} & \multirow{2}[4]{*}{mINP} \\
\cmidrule{1-2}    Depth & Width &       &       &       &       &       &  \\
    \midrule
         \Checkmark &       &       &       & 69.2  & 83.7  & 59.6  & 34.8 \\
        \Checkmark  &   \Checkmark    &       &       & 72.4  & 86.9  & 64.4  & 38.1 \\
        \Checkmark  &    \Checkmark   &     \Checkmark  &       & 74.3  & 87.4  & 65.1  & 38.7 \\
        \Checkmark  &   \Checkmark    &       &    \Checkmark   & 74.6  & 87.2  & 65.3  & 39.2 \\
        \Checkmark  &     \Checkmark  &    \Checkmark   &    \Checkmark   & \textbf{75.8}  & \textbf{88.3}  & \textbf{67.3}  & \textbf{40.4} \\
    \bottomrule
    \end{tabular}%
    }
  \label{tab:ablation}%
\end{table}%


\subsection{Visualization Analysis}

\begin{figure}[htbp]
  \centering
   \includegraphics[width=0.92\linewidth]{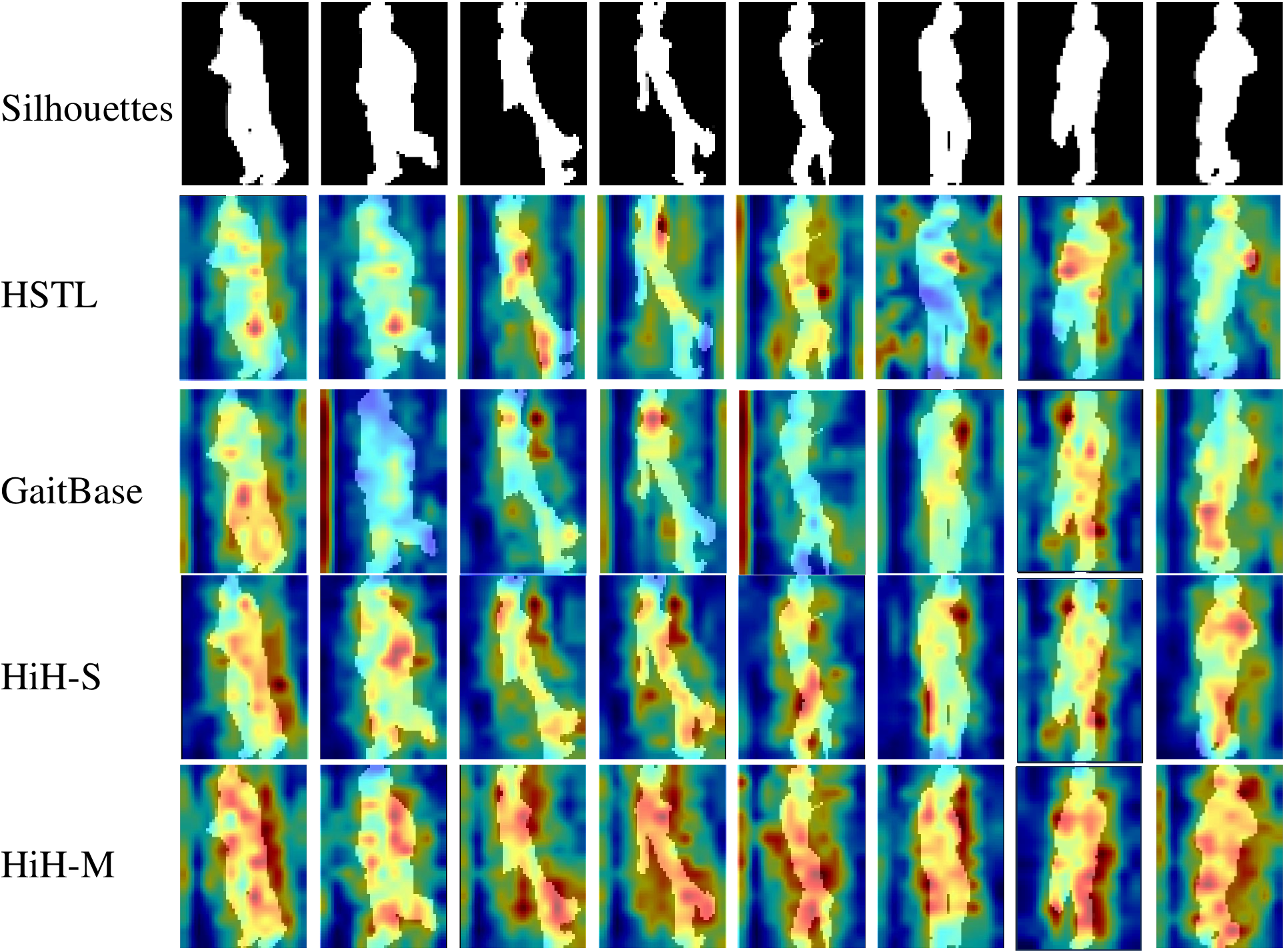}
   \caption{Comparative heatmaps of HSTL \cite{Wang_2023_ICCV}, GaitBase \cite{fan2023opengait}, HiH-S, and HiH-M on the Gait3D Dataset. }
   \label{fig:heatmap}
\end{figure}

We visualize the heatmaps from the last layer of HiH and other methods on Gait3D in \cref{fig:heatmap}. It can be observed that HSTL focuses on sparse key joints like knees, shoulders and arms, with attention on limited body parts. In comparison, GaitBase attends to more body parts within each silhouette, explaining its superior performance over HSTL. Our HiH-S further outputs denser discriminative regions across multiple views, hence achieving better results. By incorporating pose modality, HiH-M obtains more comprehensive coverage of full-body motion areas.


\subsection{Trade-off between Accuracy and Efficiency}
\label{sec:trade-off}
\begin{figure}[htbp]
  \centering
   \includegraphics[width=0.97\linewidth]{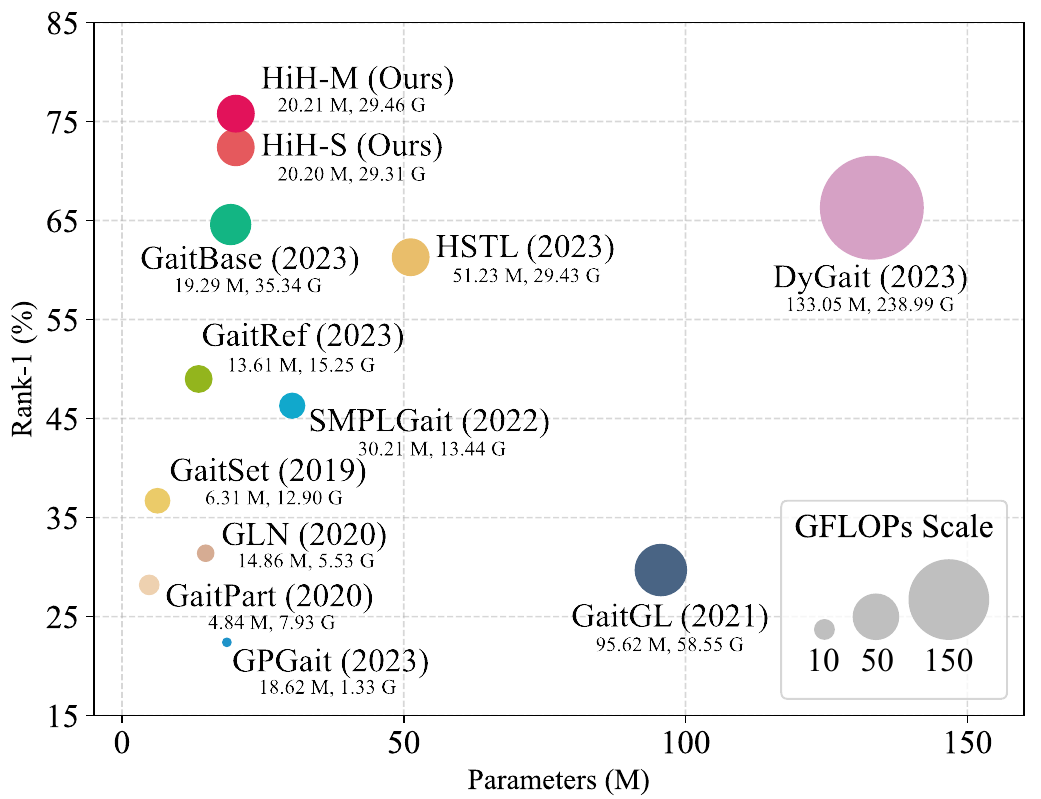}
   \caption{The trade-off between Rank-1 accuracy (\%), parameters (M), and FLOPs (G) among HIH and other methods on Gait3D.}
   \label{fig:trade-off}
\end{figure}

In our trade-off analysis, shown in \cref{fig:trade-off}, we explore the relationship between model complexity and accuracy. Pose-based methods like GPGait \cite{Fu_2023_ICCV} demonstrate parameter efficiency but fall short in performance. Clearly, larger models with higher FLOPs (Floating Point Operations per Second) generally achieve better results. DyGait \cite{Wang_2023_ICCV_DyGait}, achieving high accuracy, demands significant computation due to 3D convolutions. In contrast, GaitBase \cite{fan2023opengait} offers better parameter efficiency by utilizing 2D spatial convolutions, but incurs higher FLOPs, likely due to the lack of effective temporal aggregation mechanisms. Our HiH-S finds an optimal balance. HiH-M further enhances performance without substantially increasing computational cost. This is accomplished by using only one 2D and one 3D convolution to extract spatial and temporal dependencies from poses.

\section{Conclusion, Limitations, and Future Work}

We proposed the HiH framework, combining hierarchical decomposition with multi-modal data for multi-scale motion modeling and spatio-temporal analysis in gait recognition. While achieving state-of-the-art performance, HiH faces limitations in handling heavy occlusions and lacks automated design optimization. Future improvements for HiH include integrating 3D pose estimation from multiple views to mitigate errors, using neural architecture search for automated model design, and applying domain adaptation techniques to address challenges posed by covariates such as different clothing types.

{
    \small
    \bibliographystyle{ieeenat_fullname}
    \bibliography{main}
}

\end{document}